%% file: template.tex
\documentclass[english,fleqn]{ieej-e}
\usepackage[defaultsups]{newtxtext}
\usepackage[varg]{newtxmath}
\usepackage[superscript,nomove]{cite}
\usepackage[pdftex]{graphicx}

\usepackage{svg}
\usepackage[margin=10pt]{caption}
\usepackage[subrefformat=parens]{subcaption}

\usepackage{multirow}

\FIELD{}
\YEAR{xxxx}
\NO{1}
\title{Method of Tracking and Analysis of Fluorescent-Labeled Cells Using Automatic Thresholding and Labeling} 
\authorlist{%
 \authorentry[fukasawa@scv.cis.iwate-u.ac.jp]{Mizuki Fukasawa}{n}{IwateUniv}
 \authorentry[tomofukuda009@gmail.com]{Tomokazu Fukuda}{n}{IwateUniv}
 \authorentry[akashi@iwate-u.ac.jp]{Takuya Akashi}{m}{IwateUniv}
}
\affiliate[IwateUniv]{ Iwate University \\
4-3-5,Ueda, Morioka, Iwate, 020-8551 Japan \\ }

\begin{document}
\input{abst_and_keyword}
\maketitle

\input{chapters/introduction}

\input{chapters/cell_line}

\input{chapters/related_works}

\input{chapters/method_description}

\input{chapters/experiment_1_detail}

\input{chapters/conclusion}



\bibliography{template}

\begin{biography}
\end{biography}

\end{document}

%% file: abst_and_keyword.tex
\begin{abstract}  

High-throughput screening using cell images is an efficient method for screening new candidates for pharmaceutical drugs.
To complete the screening process, it is essential to have an efficient process for analyzing cell images.
This paper presents a new method for efficiently tracking cells and quantitatively detecting the signal ratio between cytoplasm and nuclei.
Existing methods include those that use image processing techniques and those that utilize artificial intelligence (AI).
However, these methods do not consider the correspondence of cells between images, or require a significant amount of new learning data to train AI.
Therefore, our method uses automatic thresholding and labeling algorithms to compare the position of each cell between images, and continuously measure and analyze the signal ratio of cells.
This paper describes the algorithm of our method.
Using the method, we experimented to investigate the effect of the number of opening and closing operations during the binarization process on the tracking of the cells.
Through the experiment, we determined the appropriate number of opening and closing processes.

%

%

\end{abstract}

%% file: chapters/introduction.tex
\section{Introduction}  
\label{sec:introduction}

For the efficient screening of new candidates for pharmaceutical drugs, cell image based high throughput screening is useful.
However, the efficient process of cell images is essential to finish the screening process.
In this study, we tried to automatically detect the nuclear localization process of androgen receptors with fluorescent-labeled cells.

Various analysis methods exist, including those that employ image processing techniques\cite{sato2018deterministic}. 
However, most of them can only analyze static images, and do not account for cell correspondence between images.
On the other hand, certain analysis techniques depend on artificial intelligence (AI) technology\cite{hayashida2022consistent}.
However, the applicability of the AI is limited to the type of cells on which it is trained on. 
To analyze other types of cells, a significant amount of new learning data is required.

Based on the above-mentioned points, efficient algorithms are needed to analyze the cells. 
This paper presents our developed method that allows us to quantitatively detect the signal ratio between cytoplasm and nuclei using the automatic thresholding and labeling algorithm.

%% file: chapters/cell_line.tex
\section{Cell Line}
\label{sec:cell_line}

We used immortalized human dermal papilla cells (DPCs) with the expression of R24C mutant cyclin-dependent kinase (CDK4), Cyclin D1, telomerase reverse transcriptase (TERT) \cite{fukuda2020human}.  
In addition to the immortalization, we introduced the Androgen receptor (AR) with Azami green fluorescence tag and histone H2B with monomer strawberry red fluorescence tag (Fukuda, T., et al, under submission) through the retrovirus expression.  
Based on the retrovirus transgenes, the AR can be detected by green fluorescence, and cell nuclei can be detected with red fluorescence.  We newly established cell lines for this study.

%% file: chapters/related_works.tex
\section{Related Works}
\label{sec:related_works}

There are various methods for object tracking and motion estimation.
This section describes optical flow estimation and feature detection.

\subsection{Optical Flow}
\label{subsec:optical_flow}

Optical flow refers to the patterns of apparent motion caused by relative movement between an observer and a scene. 
These patterns can be seen through the surfaces and edges of objects in the scene\cite{burton1978thinking, warren1985electronic, ROYDEN201217}.
In computer vision, optical flow refers to the distribution of apparent motion vectors of luminance patterns within an image \cite{horn1981determining}.
The technique is used to estimate motion and track objects\cite{beauchemin1995computation}.
There are two primary methods for estimating optical flow: sparse optical flow estimation \cite{lucas1981iterative} and dense optical flow estimation \cite{farneback2003two}.

Sparse optical flow estimation, such as the Lucas-Kanade method \cite{lucas1981iterative}, employs characteristic points \cite{shi1994good} on target scenes, such as corners and edges of objects, to estimate motion and track objects.
This method can provide a fast estimation by utilizing specific feature points. 
However, it may not be effective in estimating areas with few features or areas where feature points have been lost.
Specifically, for videos that record cells, the estimation may be affected by the similarity of features between each cell.
Additionally, feature points may be lost due to contact or overlap between cells during the estimation.

On the other hand, the dense optical flow estimation method, such as the Farneb\"{a}ck method \cite{farneback2003two}, utilizes information from all the pixels in a video frame to estimate motion and track objects.
Compared to the sparse approach, this method can collect more information about the motion.
However, this approach often leads to an increase in computational complexity. 
Especially for video recording cells, the dense optical flow estimation method often requires significant computational resources due to the high video resolution.

\subsection{Particle Video}
\label{subsec:particle_video}

Particle video is a motion estimation technique developed by Sand \cite{sand2008particle}.
This is an extension of dense optical flow estimation that utilizes a large number of tracking points, called particles \cite{goldman2007interactive}, to track motion.
By placing particles densely on video frames, this method accurately tracks visual movement for an extended time.
However, the use of a large number of particles in this technique results in higher computational complexity compared to dense optical flow estimation.
This is especially true for videos with high resolutions, such as those recording cells.

\subsection{Tracking Based on Feature Detection}
\label{subsec:feature_detection}

Feature detection is a technique used to identify specific areas or points in a video image.
This technique identifies key visual elements, such as corners and edges\cite{shi1994good, harris1988combined}, that dynamically change as motion occurs in the scene.
These features are then compared across multiple video frames to track objects in the scene. 
There are various feature descriptors available, including SIFT \cite{lowe2004distinctive} and SURF \cite{bay2008speeded}, which are robust to rotation, and KAZE \cite{alcantarilla2012kaze}, which is robust to local features.
However, this approach can be difficult when tracking areas with few distinctive features. 
This is especially true for videos that contain cells, where each cell may have similar features or lack distinctive features altogether.

%% file: chapters/method_description.tex
\section{Proposed Method}  
\label{sec:method_description}

\input{figures/tex/md_cell_flowchart}

The flowchart of the proposed method is shown in \mbox{Figure \ref{fig:md_cell_flowchart}}.
The method uses videos of cells which are introduced fluorescence tags.
The proposed method consists of three steps to process cells: detecting cell nuclei, tracking nuclei, and measuring the signal ratio of cells. 
To detect cell nuclei, the method uses binarization to extract the contours of nuclei. 
To track nuclei, this method uses the centroid and the bounding rectangle of the nuclei to compare the position of each nucleus in each video frame. 
To measure signal ratio, the method uses contours of nuclei to measure color information of nuclei and cytoplasm. 
By combining signal data and tracked data of cells, the method analyzes the continuous change of signal ratio of each cell.

In the best case, the proposed method can track and analyze almost 60\% of the cells in the target video.
This result is enough to detect the nuclear localization process.

\subsection{Cell Detection}  
\label{subsec:md_cell_detection}

\input{figures/tex/md_cell_detection_red_fluo}

To detect cell nuclei, the proposed method applies the process to each video frame that is captured with red fluorescence.
An example of video frames is shown in \mbox{Figure \ref{fig:md_cell_detection_red_fluo}}.

First of all, the proposed method converts the target video frame into a grayscale image.
Next, the method searches for an appropriate threshold to binarize the target image.
The procedure for automatic thresholding is explained in Section \ref{subsubsec:md_auto_thresholding}.
To detect nuclei in an image, the proposed method applies binarization to the target image using the threshold obtained through automatic thresholding. 
Then, we use a morphological opening and closing process to remove any noise present in the image. 
The number of opening and closing operations must be determined manually.
Next, we apply a labeling process to count the nuclei in the image and obtain information about their centroid and bounding rectangle. 
Finally, we assign a unique number to each nucleus.

\subsubsection{Automatic Thresholding}
\label{subsubsec:md_auto_thresholding}

To find the appropriate threshold, this method applies binarization to the grayscale image at each threshold, varying from the largest possible pixel value of the grayscale to the smallest possible value. 
Next, the opening and closing process is applied sequentially to the binarized frames in order to eliminate noise. 
Then, the method counts the number of removed noise at each threshold. 
After that, the labeling process is applied to the frames to detect and count the nuclei at each threshold. 
Then, the amount of noise is compared between thresholds, and the threshold just before the initial sharp increase in the amount of noise is determined. 
Similarly, the number of nuclei is compared between thresholds, and the threshold that detects the highest number of determined nuclei. 
Finally, the average of these two thresholds is calculated to determine the appropriate threshold value.

\subsection{Cell Tracking}  
\label{subsec:md_cell_tracking}

\input{figures/tex/md_cell_tracking_explain}

The proposed method tracks the movement of each nucleus between video frames by using the information on the shape of each nucleus obtained by cell detection. 
In the first step, the algorithm selects a target nucleus from the current frame. 
Then, the proposed method searches for the nucleus in the previous frame with a bounding rectangle enclosing the coordinate of the centroid of the target nucleus in the current frame. 
This process is visualized in \mbox{Figure \ref{fig:md_cell_tracking_explain_1}}.
After that, the method searches for a nucleus in the previous frame that has its centroid within the bounding rectangle of the target nucleus in the current frame (as shown in \mbox{Figure \ref{fig:md_cell_tracking_explain_2}}.) 
If the proposed method finds only one nucleus in the previous frame in these two searches, we associate the target nucleus and the nucleus in the previous frame as the same cell nucleus. 
If there is more than one or no nucleus, the target nucleus is excluded from the search in subsequent frames. 
These processes are applied to all nuclei in the target frame.

\subsection{Signal ratio Measurement}  
\label{subsec:md_signal_measurement}

\input{figures/tex/md_cell_detection_green_fluo}

\input{figures/tex/md_signal_measurement_mask}

To measure the signal ratio, the proposed method applies processes to each video frame that is captured with green fluorescence. 
An example of such video frames is shown in \mbox{Figure \ref{fig:md_cell_detection_green_fluo}}.
This process is applied to all tracked cells in the target frame.
First, the algorithm selects a target cell from the current frame.
Then, the proposed method acquires the target cell image from the target frame by utilizing the bounding rectangle of the nucleus. 
Next, using the binarized image obtained during the cell-detecting process, the mask image of the nucleus is created. 
Then, the mask image of the cytoplasm is created by expanding the nucleus mask image through morphological dilation and then removing the nucleus area from the expanded image.
\mbox{Figure \ref{fig:md_signal_measurement_mask}} shows examples of the mask image.
Using these two mask images, the nucleus area and cytoplasm area are extracted from the target cell image to calculate the signal ratio. 
The signal ratio is calculated by averaging the pixel values of each grayscale-converted masked image, and then subtracting the average pixel value of the cytoplasm area from that of the nucleus area. 
These processes are applied to all cells in the target frame.

%% file: figures/tex/md_cell_flowchart.tex
\begin{figure}[tb]
  \centering
  \includegraphics[width=0.4\hsize,pagebox=cropbox,clip]{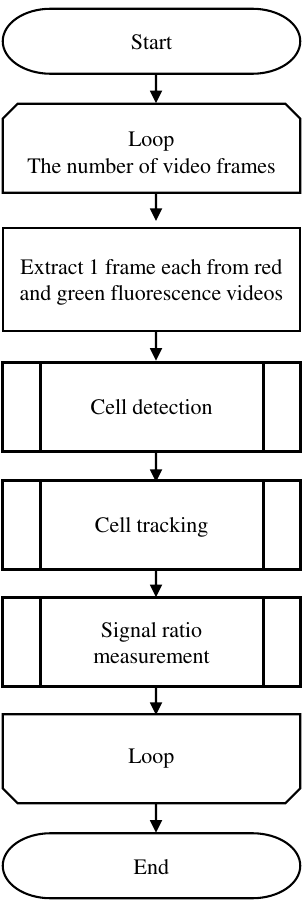}
  \caption{Flowchart of proposed method}
  \label{fig:md_cell_flowchart}
\end{figure}

%% file: figures/tex/md_cell_detection_red_fluo.tex
\begin{figure}[tb]
  \centering
  \includegraphics[width=0.8\hsize,pagebox=cropbox,clip]{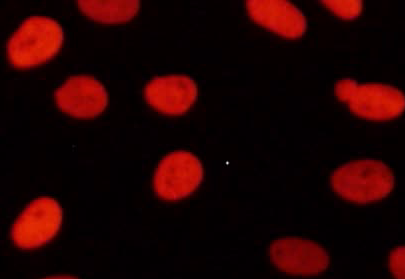}
  \caption{Video frame with red fluorescence}
  \label{fig:md_cell_detection_red_fluo}
\end{figure}

%% file: figures/tex/md_cell_tracking_explain.tex
\begin{figure*}[tb]
  \centering
  \begin{minipage}{0.3\hsize}
    \raggedright\scriptsize
    \fbox{\includegraphics[width=0.9\hsize,pagebox=cropbox,clip]{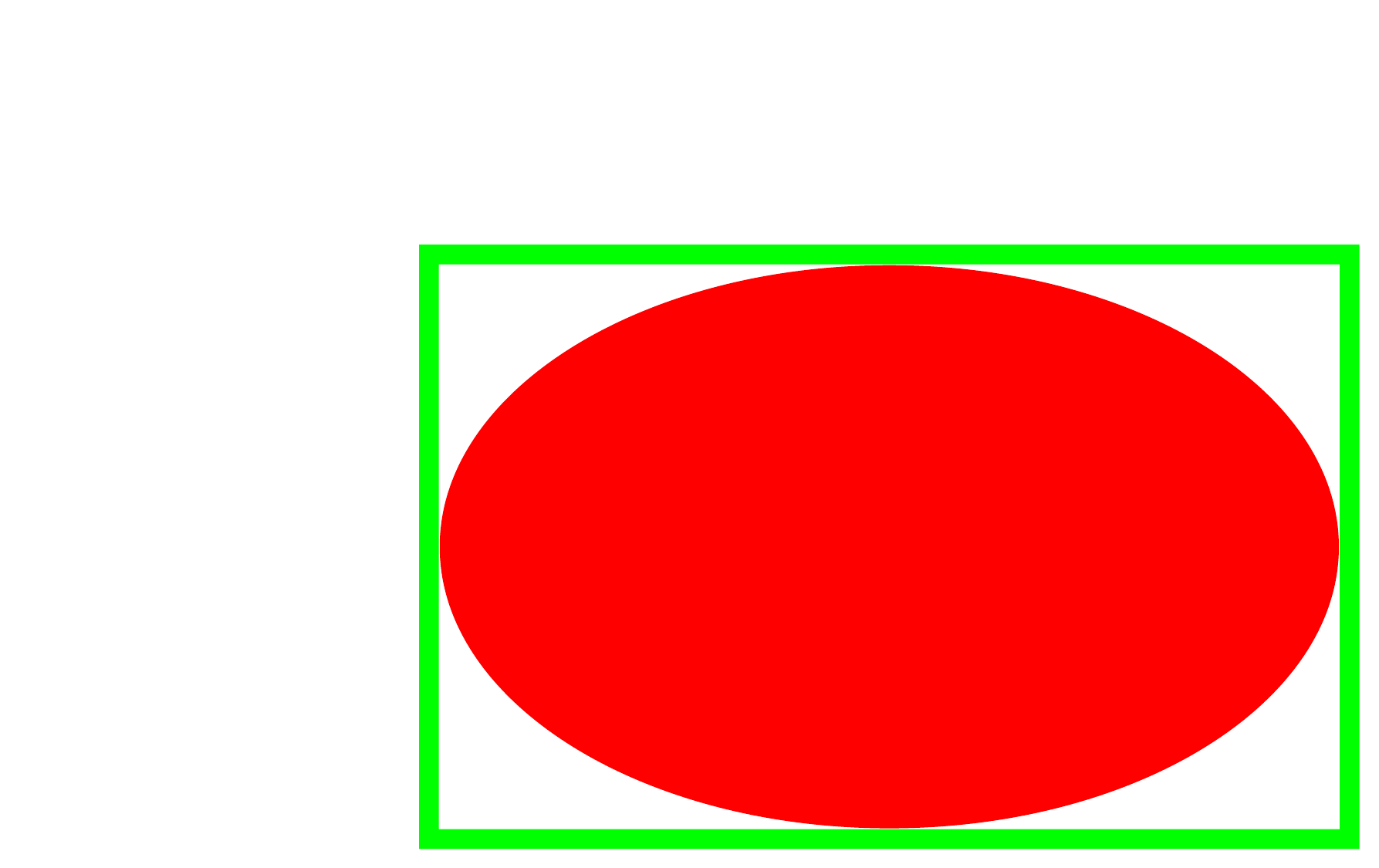}}
	\subcaption{Bounding rectangle of a nucleus in previous frame}
    \label{subfig:md_cell_tracking_prev_rect}
  \end{minipage}
  \begin{minipage}{0.3\hsize}
    \centering\scriptsize
    \fbox{\includegraphics[width=0.9\hsize,pagebox=cropbox,clip]{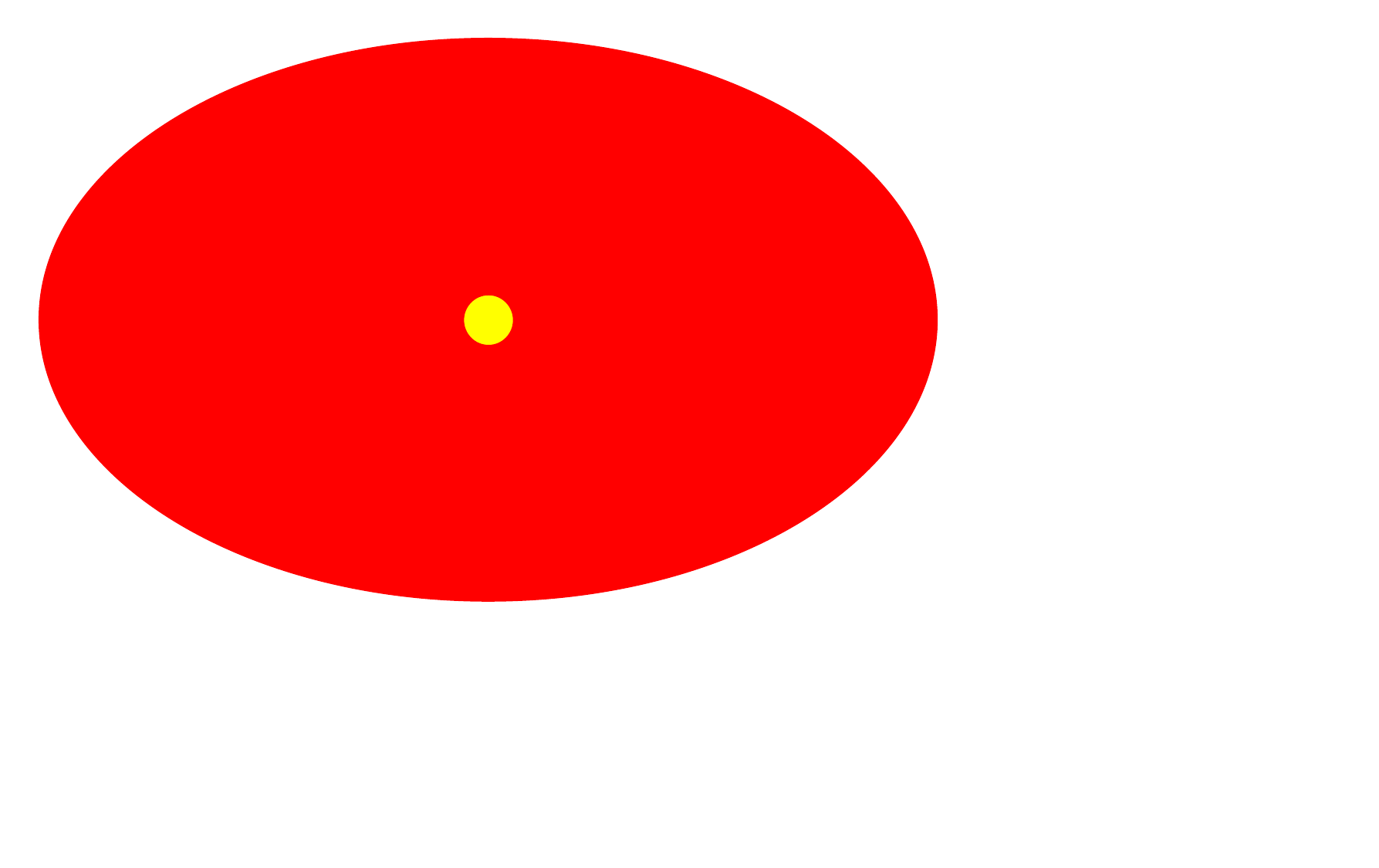}}
	\subcaption{Centroid of a nucleus in current frame\\\thickspace}
    \label{subfig:md_cell_tracking_curr_cent}
  \end{minipage}
  \begin{minipage}{0.3\hsize}
    \raggedleft\scriptsize
    \fbox{\includegraphics[width=0.9\hsize,pagebox=cropbox,clip]{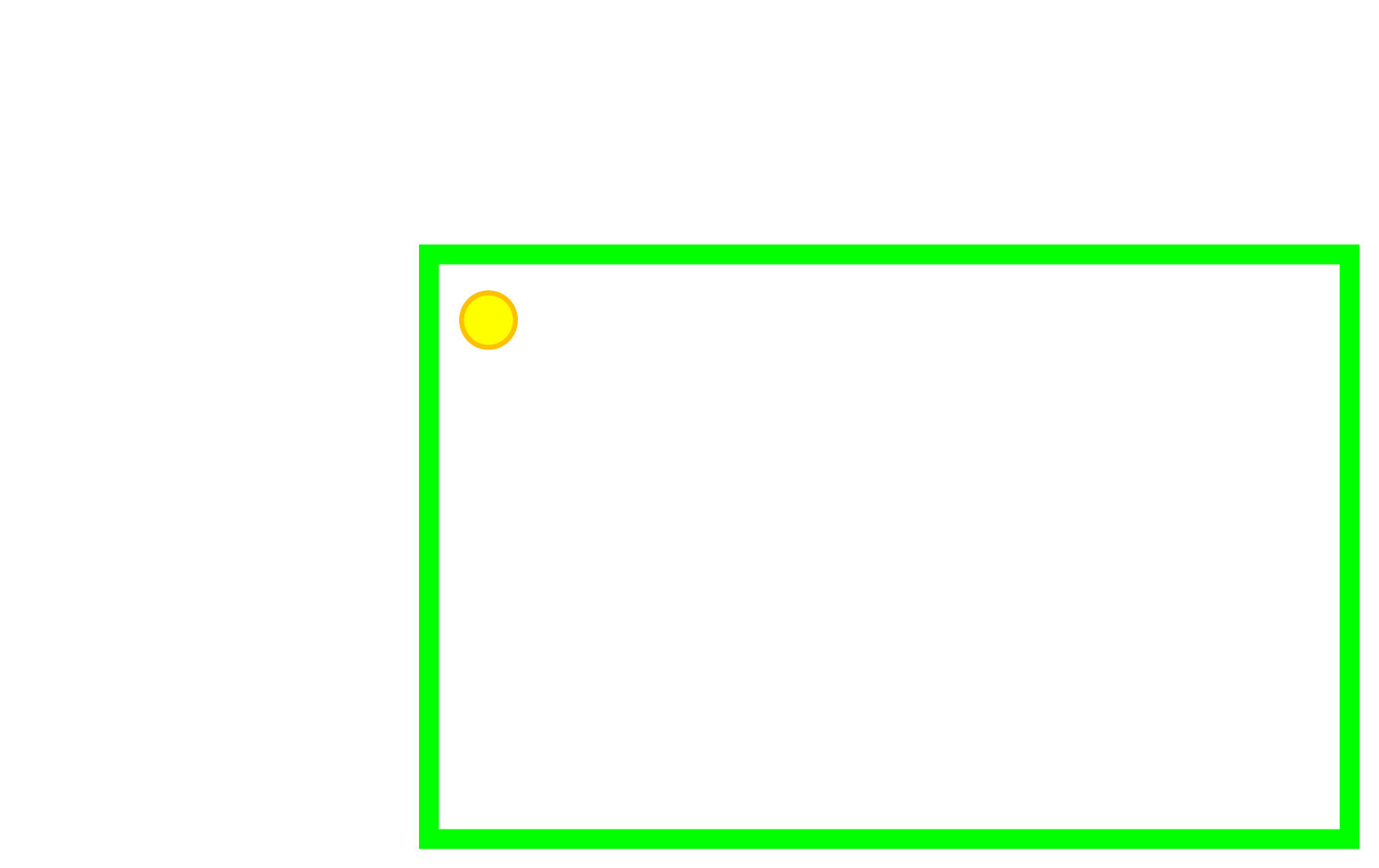}}
	\subcaption{Comparing positions of bounding rectangle and centroid}
    \label{subfig:md_cell_tracking_comp_1}
  \end{minipage}
  \caption{Process of comparing positions of bounding rectangle of previous frame and centroid of current frame}
  \label{fig:md_cell_tracking_explain_1}
\end{figure*}

\begin{figure*}[tb]
  \centering
  \begin{minipage}{0.3\hsize}
    \raggedright\scriptsize
    \fbox{\includegraphics[width=0.9\hsize,pagebox=cropbox,clip]{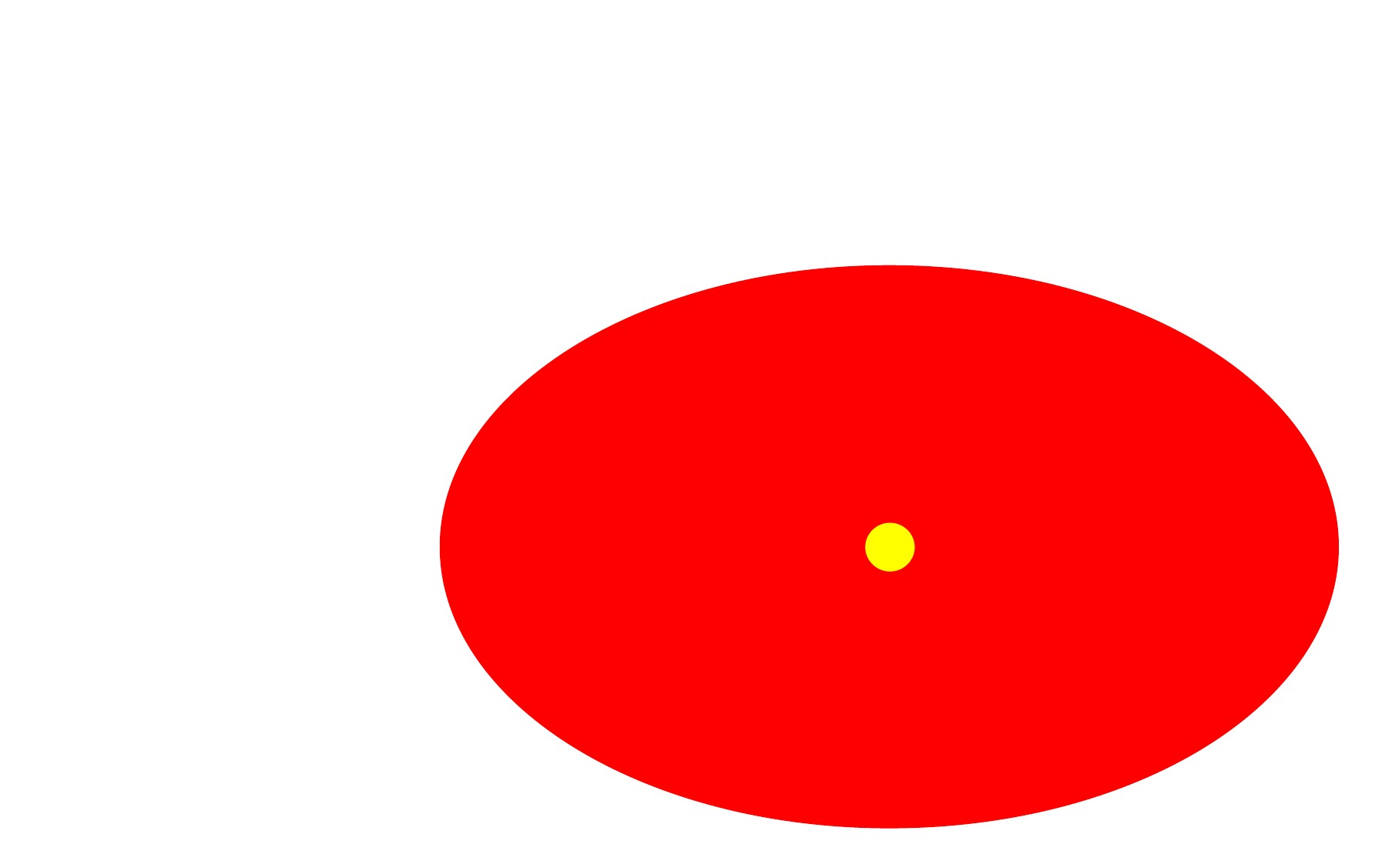}}
	\subcaption{Centroid of a nucleus in previous frame\\\thickspace}
    \label{subfig:md_cell_tracking_prev_cent}
  \end{minipage}
  \begin{minipage}{0.3\hsize}
    \centering\scriptsize
    \fbox{\includegraphics[width=0.9\hsize,pagebox=cropbox,clip]{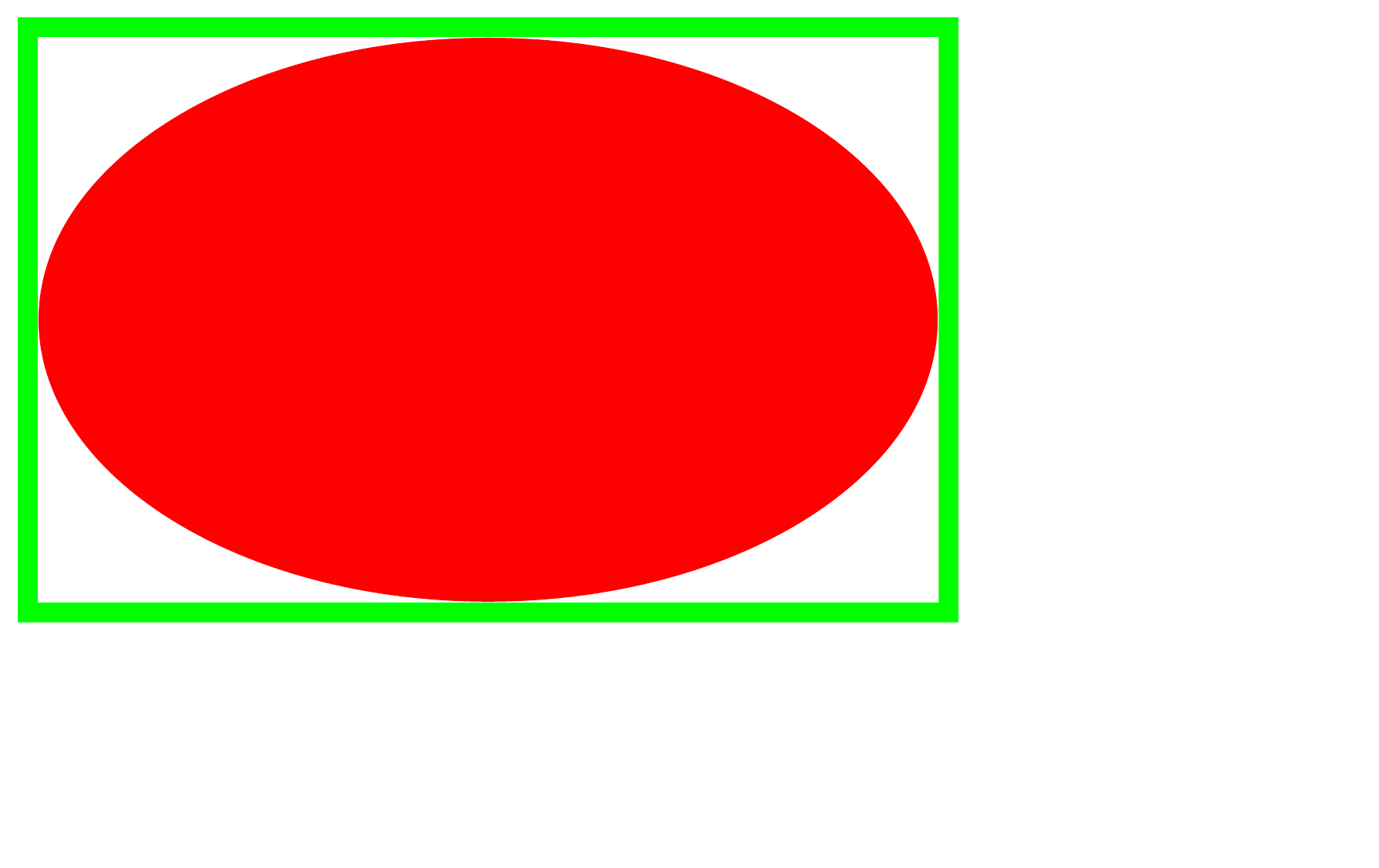}}
	\subcaption{Bounding rectangle of a nucleus in current frame}
    \label{subfig:md_cell_tracking_curr_rect}
  \end{minipage}
  \begin{minipage}{0.3\hsize}
    \raggedleft\scriptsize
    \fbox{\includegraphics[width=0.9\hsize,pagebox=cropbox,clip]{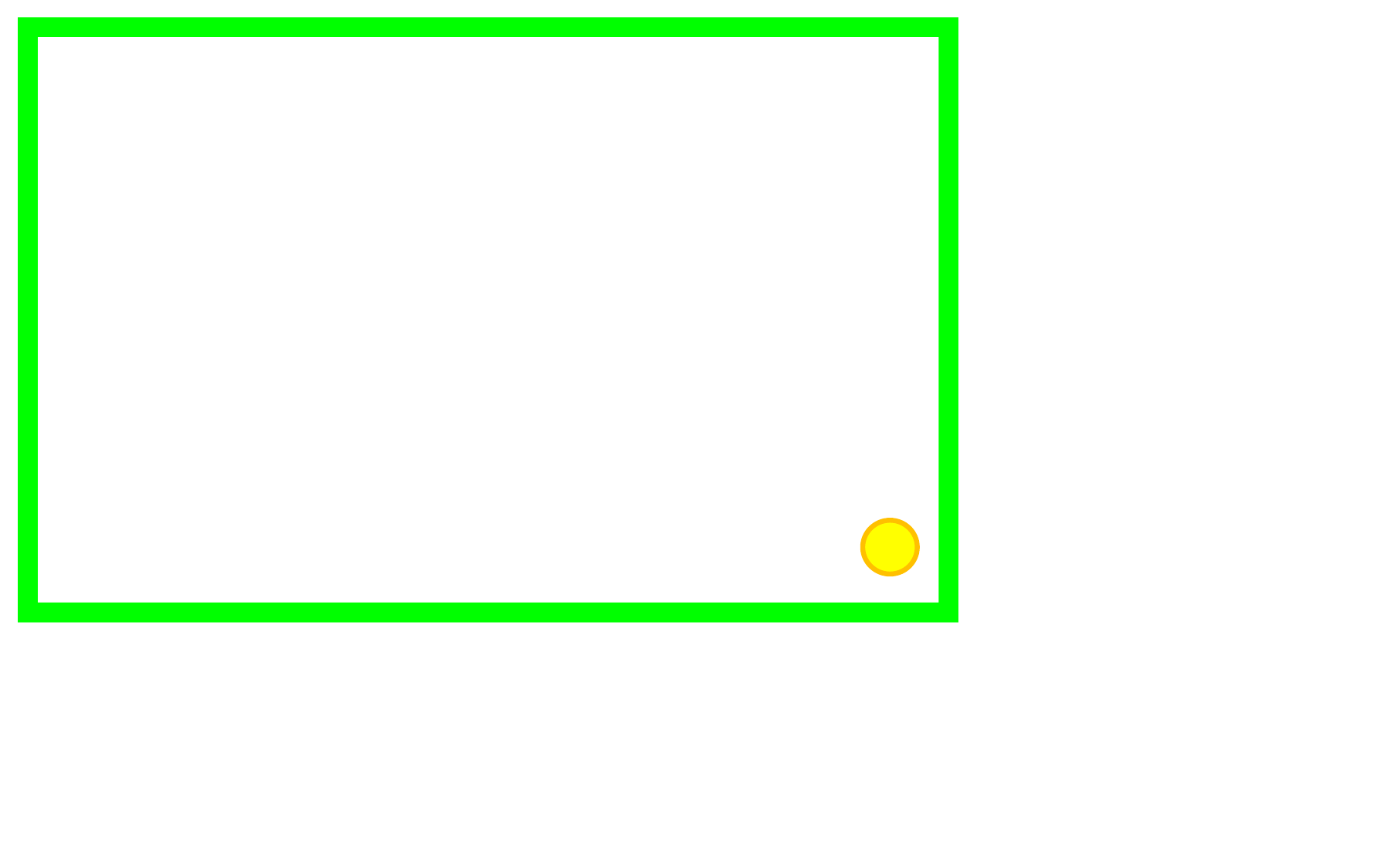}}
	\subcaption{Comparing positions of centroid and bounding rectangle}
    \label{subfig:md_cell_tracking_comp_2}
  \end{minipage}
  \caption{Process of comparing positions of centroid of a previous frame and bounding rectangle of current frame}
  \label{fig:md_cell_tracking_explain_2}
\end{figure*}

%% file: figures/tex/md_cell_detection_green_fluo.tex
\begin{figure}[tb]
  \centering
  \includegraphics[width=0.8\hsize,pagebox=cropbox,clip]{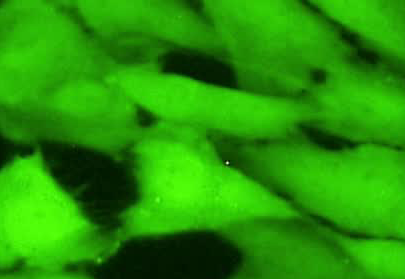}
  \caption{Video frame with green fluorescence}
  \label{fig:md_cell_detection_green_fluo}
\end{figure}

%% file: figures/tex/md_signal_measurement_mask.tex
\begin{figure}[tb]
  \centering
  \begin{minipage}{0.45\hsize}
    \raggedright\scriptsize
    \includegraphics[width=0.9\hsize]{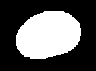}
	\subcaption{Mask image of nucleus}
    \label{subfig:md_signal_measurement_mask_nucleus}
  \end{minipage}
  \begin{minipage}{0.45\hsize}
    \raggedleft\scriptsize
    \includegraphics[width=0.9\hsize]{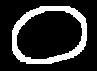}
	\subcaption{Mask image of cytoplasm}
    \label{subfig:md_signal_measurement_mask_cytoplasm}
  \end{minipage}
  \caption{Examples of mask image}
  \label{fig:md_signal_measurement_mask}
\end{figure}

%% file: chapters/experiment_1_detail.tex
\section{Experiment} 
\label{sec:experiment}

As mentioned in \mbox{Section \ref{sec:method_description}}, the proposed method can analyze enough number of cells to detect the nuclear localization process.
However, for better performance, the method needs the appropriate number of opening and closing processes.
Therefore, we conducted an experiment to investigate the effect of the number of opening and closing processes during binarization on cell tracking.

The experiment applied the proposed method to the target video while varying the number of opening and closing processes from 0 to 5. 
Then, the number of cell nuclei tracked up to the final frame is counted and compared.

\subsection{Data}  
\label{subsec:exp_data}

The analyzed video data was created from time-lapse images of a fluorescent-labeled cell population.  
Images with green fluorescence are captured to detect the AR, while images with red fluorescence are captured to detect cell nuclei.
The time-lapse images were captured at 40-second intervals using a fluorescence microscope, and later converted into videos at 15 frames per second. 
The images of the cell population with red and green fluorescence were captured simultaneously.
The length of the target videos is 16 seconds and the resolution is $1920 \times 1080$ pixels.

\subsection{Result and Discussion}  
\label{subsec:exp_result}

\mbox{Table \ref{tab:exp_result}} shows the correlation between the number of cell nuclei tracked to the final frame and the number of opening and closing processes. 
The table indicates that the number of tracked nuclei increased when the opening process was applied at least twice. 
However, the number did not increase significantly even when the opening process was applied more than twice. 
This suggests that most of the noise can be removed by applying the opening process twice.

On the other hand, as the number of closing process applications increased, the number of tracked nuclei decreased. 
This is because the closing process sometimes combines other nucleus areas, presenting them as a single nucleus. 
The proposed method considers the tracking of these nuclei as failed, and excludes them from the search.

The result became extremely small when the opening and closing processes were not applied. 
The reason for this is that the proposed method removes the tracking of nuclei from the search when they are affected by noise or overlap with other nuclei. 
Therefore, even if more targets are detected in the initial frame without any preprocessing, as the tracking progresses to subsequent frames, more targets are affected by noise and are eliminated from the tracking process.

The results obtained by applying both opening and closing processes were either equal to or less than the results obtained by only applying the opening process.
This may be because the opening process can eliminate a significant amount of noise, while the closing process sometimes combines multiple distinct nucleus areas.

In addition, repeating the opening and closing process causes deformation of the contour of nucleus masks.
The proposed method for measuring the signal ratio of cells requires information about the contour of the nuclei. 
Therefore, the proposed method has to remove noise while preserving the contour of the nuclei as accurately as possible.

\mbox{Figure \ref{fig:exp_result_deformed_cell}} presents examples of images where opening and closing processes are applied. 
As seen in \mbox{Figure \ref{subfig:exp_result_deformed_cell_5opcl}}, the processes may deform the contours of the nucleus, leading to connections with other contours.

Based on these results, it was determined that using 2 opening processes and 0 or 1 closing process is appropriate for this method.

\input{figures/tex/tab_exp_result}

\input{figures/tex/exp_result_deformed_cell}

%% file: figures/tex/tab_exp_result.tex
\begin{table}[tb]
    \caption{Correlation between the number of cell nuclei tracked to the final frame and the number of opening and closing processes}
    \label{tab:exp_result}
    \centering
    \begin{tabular}{cc||cccccc}
        \hline
        &&\multicolumn{6}{c}{Opening}\\
        && 0 & 1 & 2 & 3 & 4 & 5 \\
        \hline \hline
        \multirow{6}{*}{Closing} & 0 & 9 & 34 & 33 & 34 & 34 & 34 \\
        & 1 & 30 & 30 & 34 & 34 & 34 & 34 \\
        & 2 & 23 & 26 & 32 & 32 & 32 & 34 \\
        & 3 & 18 & 22 & 29 & 29 & 29 & 28 \\
        & 4 & 17 & 21 & 25 & 26 & 25 & 26 \\
        & 5 & 13 & 19 & 25 & 25 & 25 & 25 \\
        \hline
    \end{tabular}
\end{table}

%% file: figures/tex/exp_result_deformed_cell.tex
\begin{figure*}[tb]
  \centering
  \begin{minipage}{0.3\hsize}
    \raggedright\scriptsize
    \includegraphics[width=0.9\hsize]{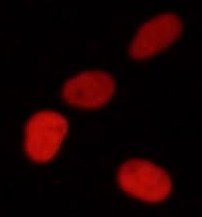}
	\subcaption{Target frame image with red fluorescence}
    \label{subfig:exp_result_deformed_cell_redfluo}
  \end{minipage}
  \begin{minipage}{0.3\hsize}
    \centering\scriptsize
    \includegraphics[width=0.9\hsize]{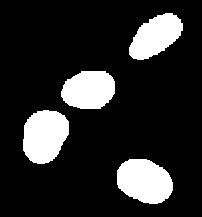}
	\subcaption{Binarized image with opening process applied twice}
    \label{subfig:exp_result_deformed_cell_2op}
  \end{minipage}
  \begin{minipage}{0.3\hsize}
    \raggedleft\scriptsize
    \includegraphics[width=0.9\hsize]{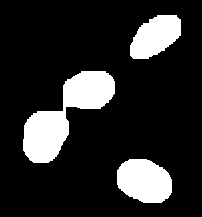}
	\subcaption{Binarized image with opening and closing process applied 5 times each}
    \label{subfig:exp_result_deformed_cell_5opcl}
  \end{minipage}
  \caption{Example of binarized image with nucleus contours that deformed or connected with other contours by opening and closing processes}
  \label{fig:exp_result_deformed_cell}
\end{figure*}

%% file: chapters/conclusion.tex
\section{Conclusion and Future Works} 
\label{sec:conclusion}

In this paper, we have proposed a method for cell tracking and signal analysis using image processing. 
The proposed method involves searching for an appropriate threshold for binarization in each video frame, and accurately tracking each cell nucleus using information about its shape.
Our method can efficiently and continuously measure the signal ratio between the cytoplasm and nuclei using the tracked cell data. 
In the experiment, we investigated the correlation between the number of tracked nuclei and the number of opening and closing processes during binarization, and determined the appropriate number of processes. 
Before submitting this paper to the journal, we will prepare the ground truth data, compare our method with others, and examine its efficiency through experiments.